\pdfoutput=1

\documentclass[11pt]{article}

\usepackage[preprint]{acl}

\usepackage{amsmath,amsfonts,bm}

\def\eqref#1{equation~\ref{#1}}

\def\1{\bm{1}}

\def\rc{{\textnormal{c}}}

\def\rvb{{\mathbf{b}}}

\def\rvp{{\mathbf{p}}}

\def\rvs{{\mathbf{s}}}

\def\rvy{{\mathbf{y}}}

\def\rmC{{\mathbf{C}}}

\def\ermC{{\textnormal{C}}}

\def\vb{{\bm{b}}}

\def\evb{{b}}

\def\evz{{z}}

\def\mC{{\bm{C}}}

\DeclareMathAlphabet{\mathsfit}{\encodingdefault}{\sfdefault}{m}{sl}
\SetMathAlphabet{\mathsfit}{bold}{\encodingdefault}{\sfdefault}{bx}{n}

\def\emC{{C}}

\newcommand{\R}{\mathbb{R}}

\newcommand{\reg}{\lambda}

\newcommand{\softmax}{\mathrm{softmax}}

\usepackage{bbm}
\def\ind{\mathbbm{1}}
\def\Pr{\mathbb{P}}

\usepackage{times}
\usepackage{latexsym}
\usepackage{booktabs}
\usepackage{xcolor}
\usepackage{tcolorbox}
\usepackage{algorithm}
\usepackage{algpseudocode}
\usepackage{nicefrac}
\usepackage{color, colortbl}
\definecolor{Gray}{gray}{0.9}

\usepackage[T1]{fontenc}

\usepackage[utf8]{inputenc}

\usepackage{microtype}

\usepackage{inconsolata}

\usepackage{graphicx}

\title{SkillAggregation: Reference-free LLM-Dependent Aggregation}

\author{
 \textbf{Guangzhi Sun\textsuperscript{1,3}},
 \textbf{Anmol Kagrecha\textsuperscript{2}},
 \textbf{Potsawee Manakul\textsuperscript{3}}
\\
 \textbf{Phil Woodland\textsuperscript{1}},
 \textbf{Mark Gales\textsuperscript{1}}
\\
\\
 \textsuperscript{1}Department of Engineering, University of Cambridge \\
 \textsuperscript{2}Department of Electrical Engineering, Stanford University \\
 \textsuperscript{3}SCB 10X, SCBX Group 
\\
 \texttt{\{gs534,pw117,mjfg100\}@cam.ac.uk, anmolk@stanford.edu, potsawee@scb10x.com}
}

\begin{document}
\maketitle
\begin{abstract}
Large Language Models (LLMs) are increasingly used to assess NLP tasks due to their ability to generate human-like judgments.
Single LLMs were used initially, however, recent work suggests using multiple LLMs as judges yields improved performance.
An important step in exploiting multiple judgements is the combination stage, aggregation. Existing methods in NLP either assign equal weight to all LLM judgments or are designed for specific tasks such as hallucination detection.
This work focuses on aggregating predictions from multiple systems where no reference labels are available. A new method called SkillAggregation is proposed, which learns to combine estimates from LLM judges without needing additional data or ground truth. It extends the Crowdlayer aggregation method, developed for image classification, to exploit the judge estimates during inference.
The approach is compared to a range of standard aggregation methods on HaluEval-Dialogue, TruthfulQA and Chatbot Arena tasks. SkillAggregation outperforms Crowdlayer on all tasks, and yields the best performance over all approaches on the majority of tasks.


\end{abstract}
\section{Introduction}


Human evaluation has long been considered the gold standard for evaluating the quality of natural language generation (NLG) systems \cite{belz-reiter-2006-comparing, lai-tetreault-2018-discourse, fabbri2021summeval}. However, human evaluation can be labour-intensive and time-consuming, especially as the complexity of language generation increases. With the advent of instruction-following large language models (LLMs) \cite{wei2022finetuned, ouyang2022training}, there has been a shift towards leveraging these models' zero-shot capabilities to evaluate NLP tasks, including NLG evaluation. Recent advancements have demonstrated high alignment between ``strong'' LLMs and human judgments across various NLP tasks \cite{zheng2023_llm_judge}. This zero-shot LLM-based method, also termed LLM-as-a-judge, offers a more cost-effective alternative to traditional human evaluation \cite{li2024_arena_hard}.


Despite its advantages, LLM-as-a-judge has limitations such as \textit{self-preference bias}, where an LLM tends to favour its own responses; \textit{verbosity bias}, where some LLMs may prefer longer, more detailed responses \cite{zheng2023_llm_judge, stureborg2024_llm_biases}; or sensitivity to prompt phrasing \cite{verga2024_llm_juries}. Using a single large LLM judge not only amplifies biases but can also require high computational resources that make local evaluations impractical. Furthermore, existing aggregation approaches in NLP weigh judges equally \cite{verga2024_llm_juries, badshah2024reference} or are tailored to specific tasks \cite{sun2024crosscheckgpt, li2024_peer_rank}, resulting in the following limitations.


{First}, assigning equal weight to all judges, as done in \citet{verga2024_llm_juries, badshah2024reference}, can be suboptimal because skill levels may vary across judges and tasks. For example, GPT-4 is expected to outperform GPT-3 in most tasks and thus should be assigned a higher weight. Claude3 might excel in programming tasks, whereas GPT-4 may surpass Claude3 in reasoning tasks, suggesting that the weighting should be adapted depending on the evaluation task.
{Second}, aggregation methods, such as those proposed by \citet{sun2024crosscheckgpt}, \citet{wei2024_fewl_alg}, and \citet{li2024_peer_rank}, are designed for specific tasks like hallucination detection or ranking LLMs. However, the applicability of these methods beyond truthfulness evaluation remains uncertain. \citet{li2024_peer_rank} impose the constraint that each judge must evaluate all others.



To address the aforementioned limitations, we propose a new method, SkillAggregation, which dynamically weights LLM judges based on contextual information, such as the specific question posed to the LLM judges when generating the estimates. This method is reference-free as it learns to combine estimates from LLM judges without needing additional data or ground truth. This method is more general as it can be applied to any problem where the LLM estimates are binary or probabilistic. Unlike prior work, we do not prompt the judges for any information besides the estimates, nor does our algorithm need each judge to assess all others. 

Our contributions are as follows. We propose SkillAggregation, an aggregation method based on a reformulation of Crowdlayer~\cite{rodrigues2018_crowdlayer}. SkillAggregation improves on Crowdlayer by learning estimates at training and utilizing them at inference.
%
Moreover, SkillAggregation includes a regularization term to mitigate over-confident probabilistic estimates from LLM judges. Finally, we demonstrate SkillAggregation's effectiveness on HaluEval-Dialogue \citep{li-etal-2023-halueval}, TruthfulQA \citep{lin-etal-2022-truthfulqa}, and Chatbot Arena \citep{lmsys-chatbot-arena} datasets.



\section{Related Work}

\textbf{Aggregation Methods}: Aggregation methods have been widely studied in fields such as crowdsourcing and ensemble learning, where strategies are developed to combine predictions (also referred to as worker predictions) to enhance decision quality during training and inference. However, few aggregation methods have been applied to LLM evaluation, despite the increasing use of multiple LLMs in NLP tasks. Most existing aggregation methods focus on weighting worker contributions without accounting for contextual information \citep{zhang2016_no_context_survey, zheng2017_no_context_survey, zhang2022_survey}), though other methods introduce context-aware mechanisms \citep{jin2020_survey, zhang2022_survey}. The references mentioned in the preceding sentences point to surveys that provide broader surveys of aggregation strategies, particularly from the crowdsourcing literature. This work builds on these insights and compares representative aggregation methods within the LLM evaluation framework.

\vspace{2mm}
\noindent \textbf{LLM-based Evaluation}: Traditional evaluation methods for NLP tasks, such as BLEU for machine translation \cite{papineni-etal-2002-bleu} and ROUGE for summarization \citep{lin-2004-rouge}, have long been task-specific and reliant on manually designed metrics. However, with the advent of LLMs capable of performing tasks in a zero-shot manner, the evaluation paradigm has shifted, and LLM-as-a-judge \cite{zheng2023_llm_judge}, where LLMs are prompted with evaluation criteria, has emerged as a flexible alternative. This method has shown strong correlation with human judgments across tasks. Building on the LLM judge, \citet{verga2024_llm_juries} introduced PoLL, a multi-judge framework where each judge is assigned equal weight. This simple aggregation led to improved performance. Recent efforts, such as CrossCheckGPT \cite{sun2024crosscheckgpt}, designed a hallucination evaluation method based on a cross-model consistency idea that uses information generated from a group of LLMs. Similarly, FEWL \cite{wei2024_fewl_alg} uses answers generated from a group of LLMs along with an answer weighing mechanism for hallucination evaluation. PRD \cite{li2024_peer_rank} adopts multiple LLMs for pairwise comparison ranking.


\section{Worker Aggregation Problem}



\begin{figure}[!ht]
    \centering
    \includegraphics[width=0.8\linewidth]{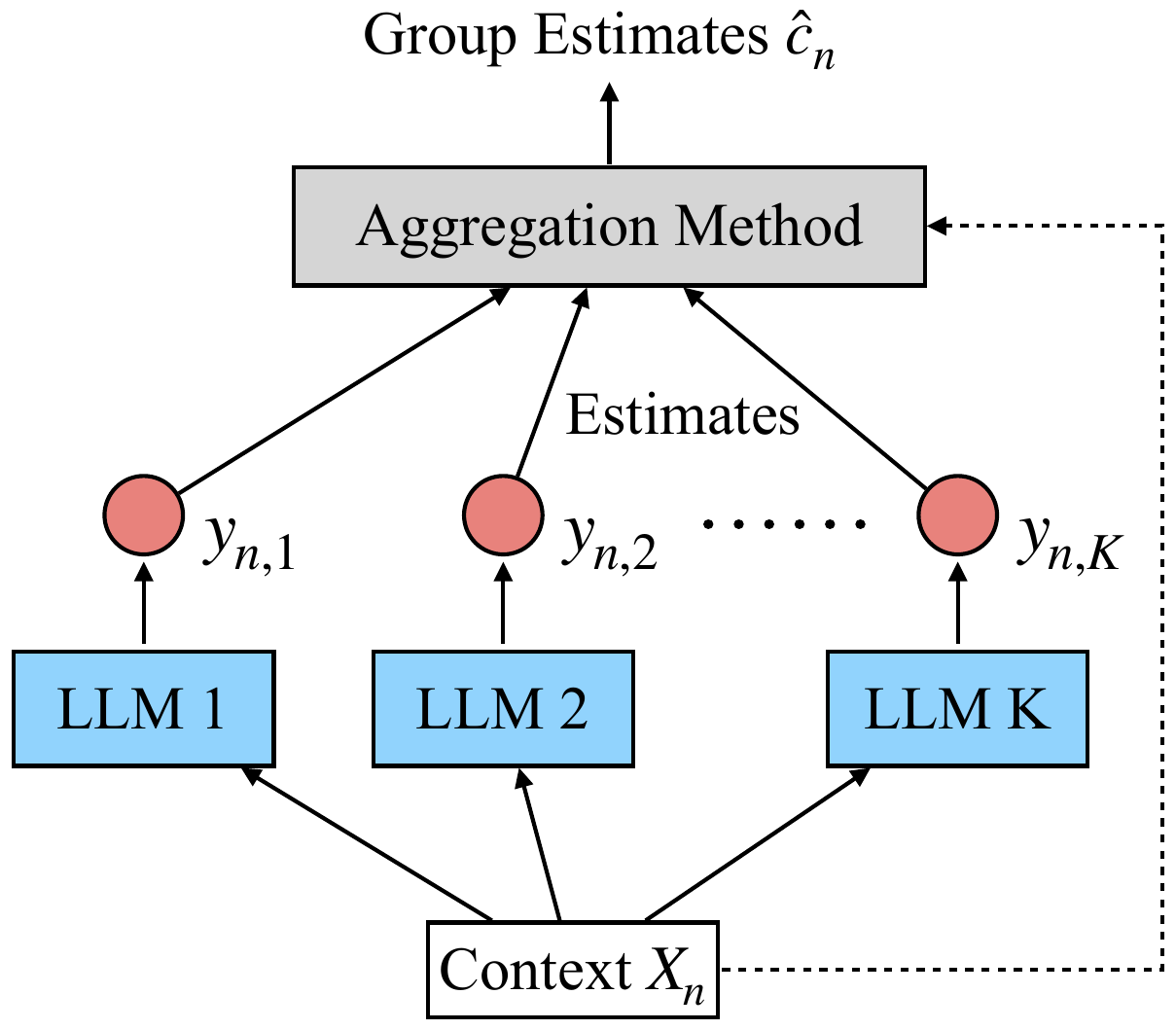}
    \caption{Illustration of the worker aggregation process. LLM judges observe context $X_n$ and provide judgments $\rvy_n \in [0,1]^K$ for ground truth $\rc_n \in \{0,1\}$. An aggregation method takes the following as input: a dataset $\{ (X_n, \rvy_n) \}_{n=1}^N$, and produces group estimates $\{\hat{c}_n \in \{0,1\}\}_{n=1}^N$ of the judgments.}
    \label{fig:prob_setting}
\end{figure}

\noindent Worker aggregation is the process of estimating the underlying ground truth using a set of predictions from a group of workers, which is crucial when dealing with multiple worker contributions that may vary in accuracy or reliability. In this paper, worker aggregation algorithms are adapted to combine predictions from LLM judges by treating each LLM as a worker to obtain a more accurate judgment. This problem is illustrated in Fig.~\ref{fig:prob_setting}.


\noindent As shown in Fig. \ref{fig:prob_setting}, there are $K$ LLM judges. For each data sample $n$, these judges observe some contextual information $X_n$ and give estimates $\rvy_n \in [0,1]^K$ for a binary ground truth label $c_n \in \{0,1 \}$ representing Yes/No or preference to A/B. Since this paper focuses on binary judgments (e.g., true/false or preference decisions) with open-source models where output probabilities can be obtained, we exploit this by having $y_{n,k}$ as the normalized probability of positive decision, e.g. ``Yes", as shown in Eqn. (\ref{eq:normprob}).
\begin{equation}
    y_{n,k} = P_n^{(k)}(\text{Yes}) / (P_n^{(k)}(\text{Yes}) + P_n^{(k)}(\text{No}))
    \label{eq:normprob}
\end{equation}
where $P_n^{(k)}(\text{Yes})$ is the $k$-th LLM output probability of generating the word ``Yes" when observing context $X_n$. We denote the binary outcomes of LLM judgments as $\rvb_n \in \{0,1 \}^K$, which have elements $b_{n,k} = \ind[y_{n,k} > 0.5]$. Note that for LLMs where probabilities cannot be obtained, binary predictions where $y_{n,k}\in\{0, 1\}$ can be used, which is a special case covered by this problem setup.




The aggregation method then combines the LLM judgments into one group estimate $\hat{c}_n$ for each sample $n$, where the input is the dataset $\{ (X_n, \rvy_n) \}_{n=1}^N$ or some representation of this dataset. Commonly adopted approaches such as averaging \cite{sun2024crosscheckgpt} and majority voting \cite{verga2024_llm_juries} are shown in Algorithm~\ref{alg:averaging} and Algorithm~\ref{alg:maj_vote} in Appendix \ref{sec:alg}, respectively.
However, both methods treat judges equally and neglect the differences in the quality and reliability of individual judges. 

On the other hand, expectation-maximization (EM) based algorithms can be applied, such as the DawidSkene algorithm \cite{dawidskene1979}. Instead of treating each judge equally, it estimates the confusion matrix associated with each judge and weights the workers based on the confusion matrices. This matrix is learned using the EM algorithm as shown in Algorithm \ref{alg:dawid_skene} in Appendix \ref{sec:alg}, and it estimates $P(b_{n,k}|c_n)$ under the assumption that contexts are not informative and $P(c_n=1)=0.5$. The group estimate $\hat{c}_n$ is taken to be the label that maximizes the approximation of $P({c}_n|\rvy_n)$.






\section{SkillAggregation}
\label{section:skill_aggregation}




\subsection{Context-dependent Aggregation}
The aforementioned aggregation methods produce the group estimates only based on the LLM judgments without considering the context that yields those judgments. With the advancement in neural networks, models have been developed that can map the context into
compact vector representations to enable further manipulation. One representative aggregation approach is the Crowdlayer \cite{rodrigues2018_crowdlayer}. Crowdlayer predicts a context-dependent distribution $\hat{P}({c}_n|X_n)$ using a neural context encoder. This context encoder, together with some transformation matrices with trainable parameters is trained to predict LLM judgments. SkillAggregation adopts a similar route with critical modifications that improve performance for aggregating LLM judgments.

\begin{figure*}[t]
    \centering
    \includegraphics[width=\linewidth]{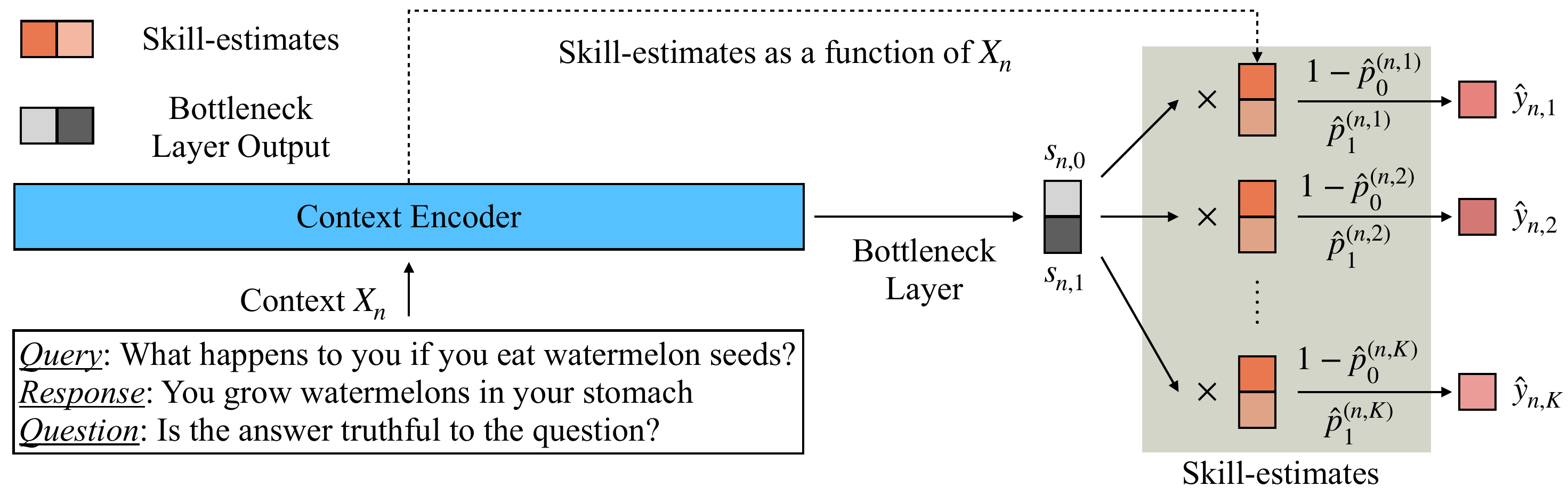}
    \caption{The SkillAggregation model with an example context from the TruthfulQA dataset. The terms $p_{0}^{(n,k)}, p_{1}^{(n,k)}$ are estimated skills for $k$-th LLM, and $s_{n,0}, s_{n,1}$ are outputs from the bottleneck layer. The outputs $\hat{y}_{n,k}$ are the predicted LLM judgments for the $k$-th LLM on data sample $n$.}
    \label{fig:skillagg_arch}
\end{figure*}

\subsection{Model Structure and Training}
\label{sec:skillmodel}

The structure of SkillAggregation is shown in Fig. \ref{fig:skillagg_arch} including a context encoder with bottleneck layer output and a set of trainable skill-estimate vectors that are defined later in this section. The context encoder is a pre-trained language model that takes the textual input and generates a vector representation of the context. Following \citet{rodrigues2018_crowdlayer}, a bottleneck layer is used to project the context representation $f_\theta(X_n)$ into a 2-dimensional vector as shown in Eqn. (\ref{eq:bottleneck})
\begin{equation}
    \rvs_n = \softmax(\mathbf{w}^Tf_\theta(X_n) + \mathbf{b})
    \label{eq:bottleneck}
\end{equation}
where $\rvs_n\in \mathbb{R}^2$ is a prediction of the distribution of the classes, i.e., $s_{n,0} \approx P(c_n=0|X_n)$. The intuition is that the bottleneck layer output would capture common information that helps predict the LLM judgments. If the estimates produced by most LLMs are better than random guessing, the output of this layer will be predictive of the ground truth.

To capture the uniqueness of each LLM judge, instead of directly learning a transformation matrix, we estimate the skills, i.e. quality and reliability of predictions from each LLM, using pairs of scalars $\hat{p}_{0}^{(n,k)} \in [0,1]$ and $\hat{p}_{1}^{(n,k)} \in [0,1]$. The underlying meaning of these two scalars is given below:
\begin{align}
    \hat{p}_{0}^{(n,k)} &\approx P(b_{n,k}=0|{c}_n=0, X_n);\\
    \hat{p}_{1}^{(n,k)} &\approx P(b_{n,k}=1|{c}_n=1, X_n).
\end{align}
We will refer to $\{\hat{\rvp}^{(n,k)}\}_{k \leq K}$ as \textit{skill-estimate} vectors since they approximate the probability of identifying the ground truth correctly. We estimate the LLM judgments using $\rvs$ and $\hat{\rvp}^{(n,k)}$ as follows:
\begin{align*}
    \hat{P}({b}_{n,k}=0|X_n) &= \hat{p}_{0}^{(n,k)}s_{n,0} + (1-\hat{p}_{1}^{(n,k)})s_{n,1}
\end{align*}
As a design choice, skill-estimate vectors can be task-specific, using the same set of $\rvp^{(n,k)}$ for all data samples $n$ in one task, or context-specific where a trainable mapping from the context to $\rvp^{(n,k)}$ is done with a linear layer followed by a Sigmoid activation function to ensure the range of values. This is referred to as \textit{SkillAggregation-X} which can be useful when the performance of LLM judges changes with subtasks or topics inside a task, allowing more flexibility.
SkillAggregation-X is trained by minimizing the cross-entropy between the predicted LLM judgments and the actual LLM judgments, as shown in Eqn. (\ref{eq:celoss})
\begin{equation}
    \mathcal{L}_\text{CE} = \sum_{n=1}^N\sum_{k=1}^K  \text{CE}(\hat{P}({b}_{n,k}=1|X_n), y_{n,k})
    \label{eq:celoss}
\end{equation}
To further analyze how skills are represented, using the fact that $s_{n,0} + s_{n,1}=1$, we re-write the estimate of $P(b_{n,k}=0|X_t)$ as follows,
\begin{align}
     &\hat{p}_{0}^{(n,k)}s_{n,0}  + (1-\hat{p}_{1}^{(n,k)})s_{n,1} \nonumber \\
    =~ &(\hat{p}_{0}^{(n,k)} + \hat{p}_{1}^{(n,k)} - 1)s_{n,0} + (1-\hat{p}_{1}^{(n,k)})
    \label{eq:linear}
\end{align}
which sets up a linear relationship between (predicted) ground truth and actual LLM judgments. An LLM with poor skill produces judgments that are less correlated with the ground truth, hence having a smaller slope. However, some LLMs are miscalibrated and give over-confident judgments. If the $k$-th LLM is over-confident which generates extreme values, slope $(\hat{p}_{0}^{(n,k)} + \hat{p}_{1}^{(n,k)} - 1)$ will be larger to minimize loss, which amplifies the influence of $y_{n,k}$ on $s_{n,0}$. This is particularly undesirable when the LLM is generating low-quality judgments. To mitigate this issue, a regularization term is proposed as shown below
\begin{equation}
    \mathcal{L}_\text{reg} = \sum_{n=1}^N\sum_{k=1}^K (\hat{p}_{0}^{(n,k)} + \hat{p}_{1}^{(n,k)} - 1)^2
    \label{eq:reg}
\end{equation}
Putting them all together, SkillAggregation is optimized by the following loss
\begin{equation}
    \label{eq:loss_fn_com}
    \mathcal{L} = \mathcal{L}_\text{CE} + \reg \mathcal{L}_\text{reg}
\end{equation}
where $\reg$ is a hyper-parameter controlling the effect of the regularization term.

\subsection{Inference with Posterior Estimation}
The Crowdlayer method performs inference by directly selecting the class with the highest probability from the bottleneck output $\rvs_n$ 
without utilizing worker predictions. Specific to our problem setup, the LLM judgments are available for each context during inference. Therefore, we propose to choose the best class based on the estimation of the posterior distribution, $P(c_n|X_n, \rvy_n)$. This is useful when the context encoder is a relatively small LM and is not powerful enough to make an accurate prediction for the ground truth, and the LLM judge predictions are, in general, better than the bottleneck outputs. Posterior estimation enables us to produce better group estimates by using powerful LLM judgments in addition to the output of the bottleneck layer.

To derive the expression for the posterior distribution, we first make an assumption common in the crowdsourcing literature that the LLMs are conditionally independent (CI) given the same ground truth and context. With this assumption, the probability of observing the set of binary LLM judgments $\vb \in \{0,1\}^K$ is
\begin{align*}
&P(\rvb_n = \vb | X_n, \rc_n=1)= \\
 &\qquad\qquad\prod_{k=1}^K \left(p_{1}^{(n,k)}\right)^{\evb_{k}}\left(1-p_{1}^{(n,k)}\right)^{1-\evb_{k}} \text{ and} \\
&P(\rvb_n = \vb | X_n, \rc_n=0)= \\
  &\qquad\qquad\prod_{k=1}^K \left(p_{0}^{(n,k)}\right)^{1-\evb_{k}}\left(1-p_{0}^{(n,k)}\right)^{\evb_{k}},
\end{align*}
for some $\{p_{k}^{(n,k)}\}_{k=1}^K \in [0,1]^K$ and $\{p_{1}^{(n,k)}\}_{k=1}^K \in [0,1]^K$. Note that the $\{\rvp^{(n,k)} \}_{k=1}^K$ are true \textit{skill} vectors under the CI model in contrast to the skill-estimate vectors, that is, $\{\rvp^{(n,k)} \}_{k=1}^K$ are equal to the probability of identifying the ground truth correctly. Under the CI assumption, the Bayes rule yields the following
\begin{align*}
    &P(\rc_n=1|X_n, \rvb_n) \propto \\
    & P(\rc_n=1|X_n) \prod_{k=1}^K \left(p_{1}^{(n,k)}\right)^{\evb_{k}}\left(1-p_{1}^{(n,k)}\right)^{1-\evb_{k}} \text{,} \\
    & P(\rc_n=0|X_n, \rvb_n) \propto \\
    & P(\rc_n=0|X_n) \prod_{k=1}^K \left(p_{0}^{(n,k)}\right)^{1-\evb_{k}}\left(1-p_{0}^{(n,k)}\right)^{\evb_{k}}.
\end{align*}
Next, we approximate the true skill vectors and the probability of the ground truth given context with the ones estimated using the SkillAggregation model to derive tractable implementation, that is:
\begin{align*}
    \Pr(\rc_n=i|X_n) \approx s_{n,i} \text{ and } 
    p_{i}^{(n,k)} \approx \hat{p}_{i}^{(n,k)}
\end{align*}
for $i \in \{0,1\}$. Group estimate maximizes the approximation of $P(c_n|X_n,\rvb_n)$ and the logic of producing it is given below: 
\begin{align}
    \label{eq:combine_ests_bottleneck}
    r_n &= \frac{s_{n,1} \prod_{k=1}^K \left(\hat{p}_{1}^{(n,k)}\right)^{b_{n,k}}\left(1-\hat{p}_{1}^{(n,k)}\right)^{1-b_{n,k}}}{ s_{n,0} \prod_{k=1}^K \left(\hat{p}_{0}^{(n,k)}\right)^{1-b_{n,k}}\left(1-\hat{p}_{0}^{(n,k)}\right)^{b_{n,k}}}, \nonumber \\
    \hat{\rc}_n & = \ind[r_n > 1].
\end{align}
This estimation relies on the quality of the skill-estimate vectors; hence, it also benefits from the proposed regularization term that prevents the impact of over-confident LLMs.\footnote{Extension of SkillAggregation to multi-class classification is shown in Appendix \ref{app:multiclass}.}

\section{Experimental Setup}



\subsection{Tasks \& Datasets} 

We consider three tasks to evaluate the performance of worker aggregation methods, including HaluEval-Dialogue \citep{li-etal-2023-halueval}, TruthfulQA \citep{lin-etal-2022-truthfulqa}, and Chatbot Arena \citep{lmsys-chatbot-arena}. 
HaluEval-Dialogue and TruthfulQA are question-answering tasks where the LLMs are prompted to determine if a given answer is correct or not. In Chatbot Arena, LLMs are used to predict human preferences between two responses that can be used for comparative assessments \cite{comparative} or reinforcement learning \cite{rlaif}. Specifically, the LLMs are made to determine if response A is better than response B by presenting both responses in the prompt. More details about the datasets are given in Appendix~\ref{app:datasets}.

\subsection{LLM Judges}

Our experiments are performed primarily with ten LLM judges that have 7B, e.g. Mistral-7B-Instruct-v0.2~\cite{mistral} or 8B, e.g. Llama-3-8B-Instruct~\cite{llama3} model parameters. These models are all instruction-tuned so that they can give judgments for a specific task. In addition, for TruthfulQA and Chatbot Arena, we remove judges that fail to perform these tasks, i.e. those having performance near or worse than random guessing, by evaluating a small number of development examples, as not all judges can perform all tasks. As a result, 8 judges are used for TruthfulQA and 9 judges are used for Chatbot Arena. The performance of each LLM judge on the three tasks is shown in Appendix \ref{app:perf_each_judge}.

We further examine the effect of worker aggregation methods using 70B-level LLMs (e.g. Mixtral-8x7B and Llama-3-70B-Instruct) that have much better performance than 7B/8B models. Five different models are selected for both tasks. Details about the models we consider are given in Appendix~\ref{app:perf_each_judge}, and the specific prompts we use to elicit estimates are given in Appendix~\ref{app:llm_judge_prompts}.

\subsection{Alternative Aggregation Methods}


We examine simple baselines, including majority voting and averaging, as well as alternative aggregation methods that have been widely adopted in the crowdsourcing literature but have not been explored on LLM-based evaluation tasks as follows:

\textbf{1) Averaging Probabilities}: This method averages the normalized probabilities of generating Yes/No or A/B from the LLM judges, and then makes predictions based on the averaged probabilities. See Algorithm~\ref{alg:averaging} in Appendix \ref{sec:alg} for the detailed implementation.

\textbf{2) Majority Voting}: Majority voting first converts the probabilistic judgments into binary decisions, and chooses the decision that agrees with more than half of the LLMs. See Algorithm~\ref{alg:maj_vote} in Appendix \ref{sec:alg} for the detailed implementation.

\textbf{3) Train on Majority Voting}: We train the same backbone context encoder to predict group estimates produced by majority voting based on the context. This is because the pre-trained context encoder may have an inherent ability to map the context to the target space with noisy training labels. By comparing to this system, we exclude the influence from the inherent knowledge when demonstrating the effect of aggregation methods.

\textbf{4) DawidSkene}: This was proposed by \citet{dawidskene1979}. We consider it representative of methods that do differential weighting, but the weights do \textit{not} depend on the context. It uses the expectation-maximization (EM) algorithm to learn skills for each worker. See Algorithm~\ref{alg:dawid_skene} in Appendix \ref{sec:alg} for the detailed implementation.

\textbf{5) Crowdlayer}: We refer to the Crowdlayer approach as the one proposed by \citet{rodrigues2018_crowdlayer}. We consider it representative of methods that do context-based aggregation methods using neural networks. 

We treat averaging probabilities, majority voting and training on majority voting as our baselines. DawidSkene and Crowdlayer are two representative common aggregation algorithms that are first applied to combine LLM judgments in this paper.

section{Model and Training Setup} 

Unless specified, we use pre-trained GPT-2, the base model with 117M parameters, \cite{radford2019language} as the context encoder, with the final hidden state representing the context. All aggregation methods are \textit{reference-free}, and we learn directly from LLM judgments using the entire test set with \textit{no} reference labels. To select model checkpoints, we randomly select a small development set of 250 samples with reference labels per task. This set is small enough to have negligible influence on the overall dataset, and training on it does not yield good performance across the test set. The hyper-parameter tuning for neural methods is based on the development set accuracies. On average, training SkillAggregation takes only 20-30 minutes on a single NVIDIA RTX 6000 Ada.





\begin{table*}[t]
    \centering
    \small
    \begin{tabular}{lcccccc}
    \toprule
      & \multicolumn{2}{c}{HaluEval (\%)} & \multicolumn{2}{c}{TruthfulQA (\%)} & \multicolumn{2}{c}{Chatbot Arena (\%)} \\
    &  7B/8B & $\sim$ 70B & 7B/8B & $\sim$ 70B & 7B/8B & $\sim$ 70B \\
    \midrule
    \rowcolor{Gray}\multicolumn{7}{l}{Context-Independent Methods} \\
    Average Probability & 76.28 & 81.10 & 68.06 & 83.85 & 63.24 & 70.65 \\
    Majority Voting	& 76.16 & 80.81 &  67.47 & 83.63 & 63.93 & 70.61 \\
    DawidSkene & 76.78 & 80.86 & 67.84 & 84.08 & \textbf{64.71} & 70.64 \\
    \rowcolor{Gray}\multicolumn{7}{l}{Context-Dependent Methods} \\
    Train on Majority Voting	& 78.78$_{\pm1.10}$ & 82.97$_{\pm0.73}$ & 67.32$_{\pm0.22}$ & 82.41$_{\pm0.43}$ & 63.77$_{\pm0.15}$ & 70.53$_{\pm0.21}$ \\
    Crowdlayer	& 79.27$_{\pm0.39}$ & 83.94$_{\pm0.33}$ & 67.74$_{\pm0.69}$ & 83.87$_{\pm0.29}$ & 64.06$_{\pm0.21}$ & 70.38$_{\pm0.06}$ \\
    SkillAggregation w/o. Reg. & 80.22$_{\pm0.41}$ & 84.22$_{\pm0.32}$ & 68.07$_{\pm0.30}$ & 84.04$_{\pm0.38}$ & 64.17$_{\pm0.07}$ & 70.62$_{\pm0.23}$ \\
    SkillAggregation & \underline{80.83$_{\pm0.33}$} & \textbf{84.88}$_{\pm0.21}$& \underline{68.74$_{\pm0.13}$} & \underline{84.45$_{\pm0.53}$} &  64.22$_{\pm0.04}$ & \underline{70.66$_{\pm0.07}$} \\
    SkillAggregation-X & \textbf{81.06$_{\pm0.14}$} & \underline{84.79$_{\pm0.13}$} & \textbf{68.77$_{\pm0.21}$} & \textbf{84.57$_{\pm0.17}$}  & \underline{64.43$_{\pm0.11}$} & \textbf{70.72$_{\pm0.07}$} \\
    \bottomrule
    \end{tabular}
    \caption{Judgment accuracies on HaluEval-Dialogue, TruthfulQA and Chatbot Arena datasets using context-independent and context-dependent aggregation methods. The second-best results are underlined. Standard deviations in 5 runs with different random seeds are also reported for context-dependent methods is required.}
    \label{tab:main7b}
\end{table*}

\section{Results and Analysis}

\subsection{Main Results}

The main results for 7B/8B models on HaluEval-Dialogue, TruthfulQA, and Chatbot Arena are shown in Table \ref{tab:main7b}. Differential weighting methods like DawidSkene and SkillAggregation perform better than majority voting on all three tasks. In particular, Crowdlayer-based methods consistently outperformed both majority voting as used in \citet{verga2024_llm_juries} and averaging as used in \citet{sun2024crosscheckgpt}, offering a better way to combine LLM estimates. The best performance across all neural methods is obtained using SkillAggregation-X, which achieved 4.9\%, 1.3\%, and 0.5\% absolute accuracy improvements on HaluEval-Dialogue, TruthfulQA, and Chatbot Arena, respectively. 

\textbf{Effect of the regularization term}: Consistent improvements are observed using the regularization term across all three datasets, especially with 7B/8B models. Without regularization, over-confident skill-estimate vectors are learned, dominating the posterior distribution and yielding a sub-optimal aggregation effect. Besides, context-specific skill-estimate vectors generally perform slightly better than task-specific ones because context-specific skill-estimate vectors capture subtle variations in these tasks.

\textbf{Larger improvements on HaluEval}: The improvements on HaluEval are considerably larger compared to the other two tasks, likely because the context encoder (GPT-2) already has some inherent ability to perform this task. By simply training on majority voting, a 1.5\% gain is already achieved on HaluEval compared to majority voting, however, on other datasets training on majority voting leads to worse performance. This suggests that the context is more informative, resulting in higher improvements with context-dependent methods. Compared to TruthfulQA, the context in Chatbot Arena is more challenging to encode, and human preference evaluation is inherently noisier, leading to the smallest gains across the three datasets. 

\textbf{SkillAggregation on 70B-level models}: As shown in Table~\ref{tab:main7b}, applying aggregation methods to 70B LLMs gives much better performance compared to 7B/8B LLMs. While differential weighting aggregation methods consistently outperform majority voting and SkillAggregation, achieving the best performance, the gains are smaller. This is particularly noticeable in Chatbot Arena, where the impact of noisy labels is more pronounced.



\subsection{Visualization of Skills}

\begin{figure}[h]
    \centering
    \hspace{-0.3cm}
    \includegraphics[width=0.9\linewidth]{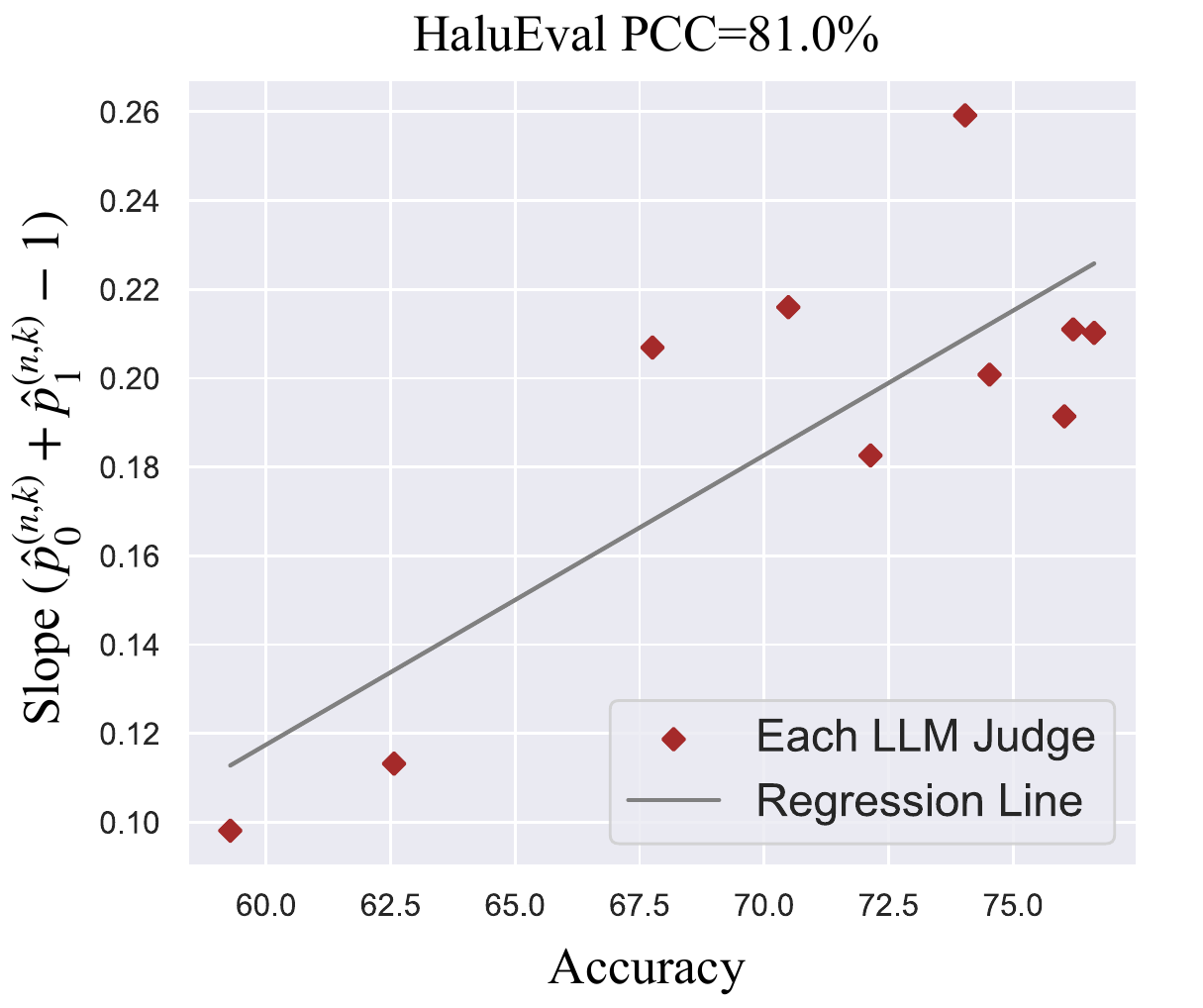}
    \caption{Scatter plot of the slope $(\hat{p}_{0}^{(n,k)} + \hat{p}_{1}^{(n,k)} - 1)$ representing the skill of each LLM from SkillAggregation against accuracies of each LLM on HaluEval. PCC stands for Pearson Correlation Coefficient.}
    \vspace{-0.5cm}
    \label{fig:skillpcc}
\end{figure}

As discussed in Section \ref{sec:skillmodel}, the learned skills of the LLMs can be reflected by the slope in Eqn.~(\ref{eq:linear}), and the slope against accuracies for each LLM judge on HaluEval is shown in Fig. \ref{fig:skillpcc}. As a result, these learned skills demonstrate a strong correlation with performance such that models with weaker abilities are under-weighed in the posterior computation during inference. Similar plots for TruthfulQA and Chatbot Arena are shown in Appendix \ref{app:skills} with PCCs of 89.9\% and 69.5\%, respectively.

\subsection{Influence of Context Encoder}

\begin{figure*}[t]
    \centering
    \includegraphics[width=0.95\linewidth]{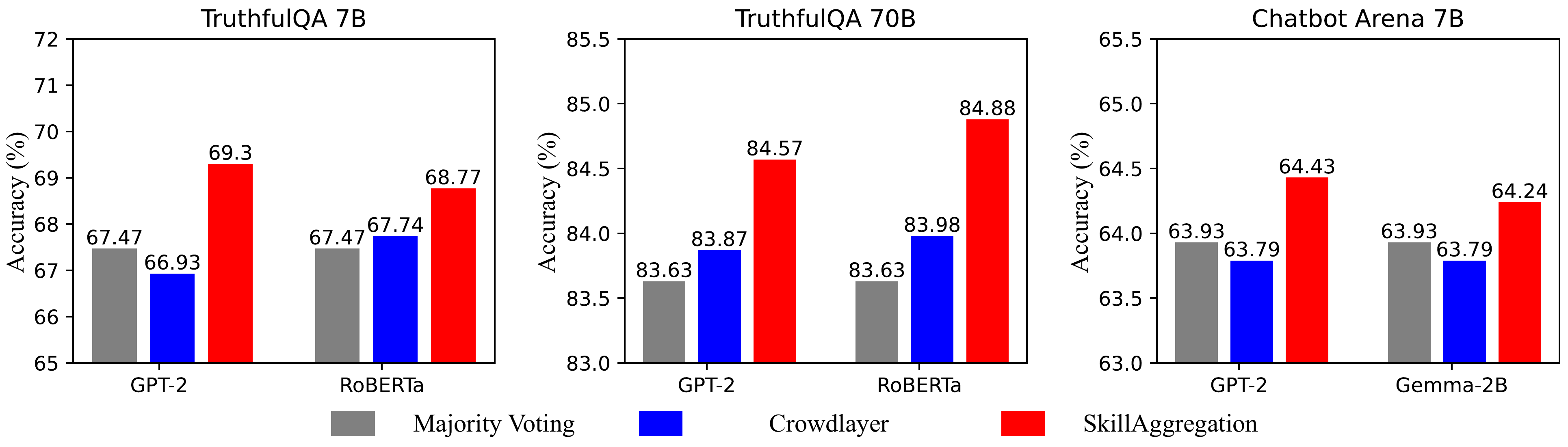}
    \caption{Accuracy when replacing the GPT-2 context encoders with RoBERTa and Gemma-2B for Crowdlayer and SkillAggregation on TruthfulQA and Chatbot Arena datasets.}
    \label{fig:backbone}
\end{figure*}

To validate that our context-dependent method is agnostic to the choice of context encoder, provided that the encoders have similar abilities, we replace the GPT-2 context encoder with pre-trained RoBERTa and use the \texttt{[CLS]} token as the context representation. For Chatbot Arena, due to longer context lengths, we chose Gemma-2B trained with Low-rank adaptation (LoRA). The results are plotted in Fig. \ref{fig:backbone}. As shown in Fig. \ref{fig:backbone}, SkillAggregation with both RoBERTa and Gemma-2B achieved similar performance improvements compared to majority voting and Crowdlayer. While Gemma-2B is larger than GPT-2, it does not necessarily give better context representations than GPT-2 for human preference prediction, hence yielding a similar performance. This is also supported by the fact that Gemma-2B achieves an accuracy of 85.59\% on TruthfulQA in contrast to 84.57\% with GPT-2.

\subsection{Subsets of LLM Judges}

\begin{figure}[!ht]
    \centering
    \includegraphics[width=0.85\linewidth]{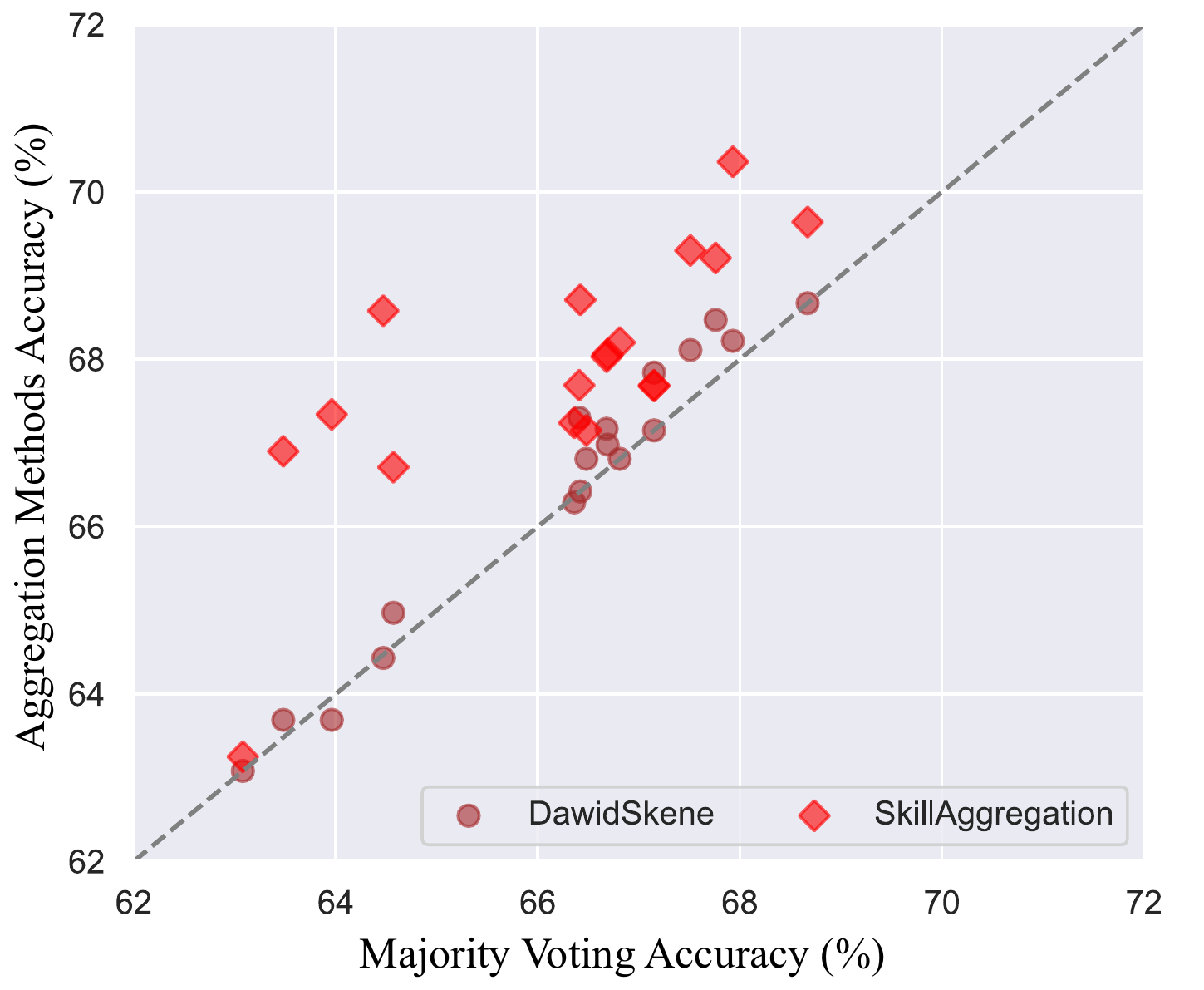}
    \caption{Accuracies of SkillAggregation and DawidSkene against majority voting on different subsets of LLM judges on TruthfulQA. The equal-performance line is plotted, and points on the upper-left side indicate improved performance compared to majority voting.}
    \label{fig:subsets}
\end{figure}

Here, we investigate the impact of LLM judge subsets on the performance of DawidSkene and SkillAggregation, with accuracies compared to majority voting in Fig.~\ref{fig:subsets}. On most subsets, SkillAggregation outperformed DawidSkene and majority voting. The majority voting performance indicates the quality of the judges in the subset. When judges are weak, SkillAggregation performs close to DawidSkene, which does not use context. In this scenario,  SkillAggregation cannot learn a good estimate of the ground truth distribution, i.e., $P({c}_n|X_n)$. On the contrary, SkillAggregation shows larger improvements when most judges perform above average and when there are a few weak models that deteriorate the majority voting performance, such as when majority voting accuracy is around 65\%-68\% in Fig. \ref{fig:subsets}. Performance on Chatbot Arena is provided in Appendix \ref{app:subsets2}.

\subsection{Influence of Dataset Sizes}

\begin{figure}[h]
    \centering
    \includegraphics[width=1.0\linewidth]{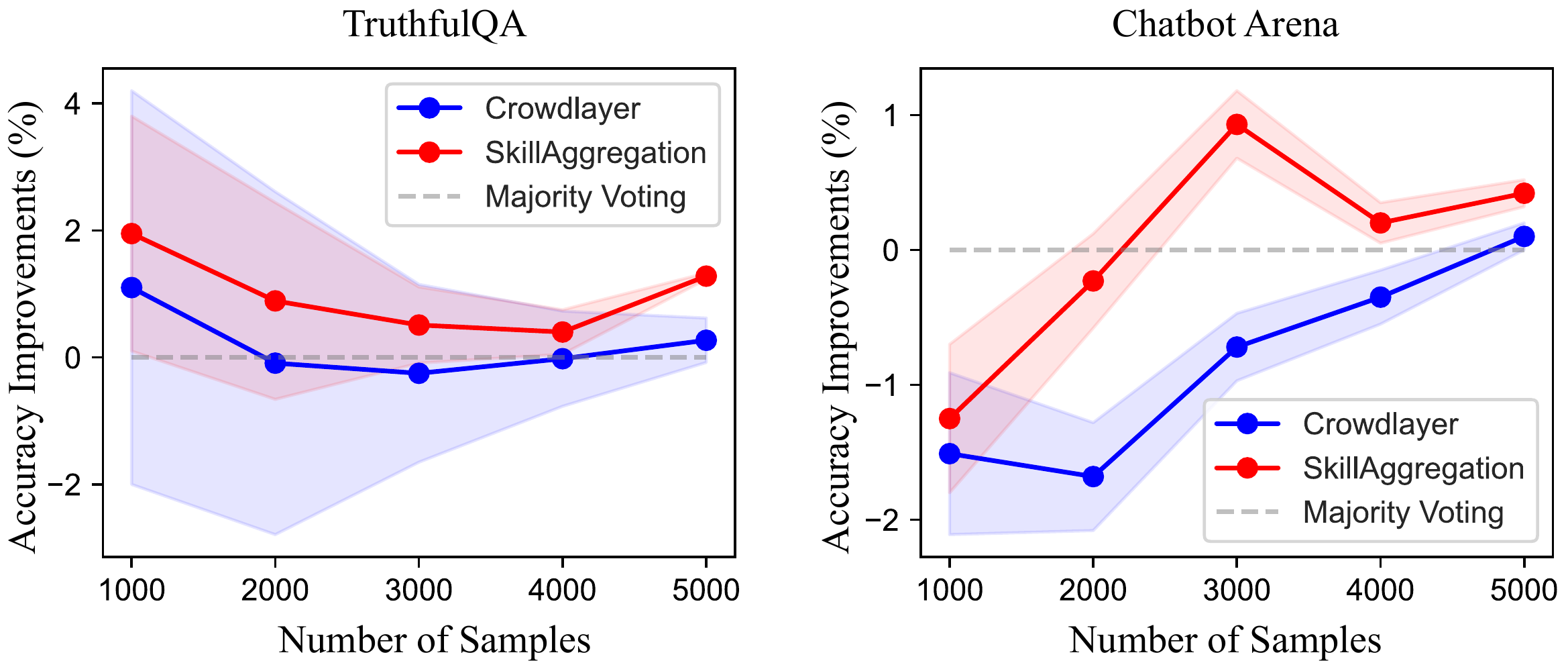}
    \caption{Accuracy improvements relative to majority voting against number of samples on TruthfulQA and Chatbot Arena using Crowdlayer and SkillAggregation, where 5000 samples for TruthfulQA is the entire dataset, and is 1/6 of Chatbot Arena.}
    \label{fig:size_diff}
\end{figure}

The impact of dataset size on SkillAggregation's performance is shown in Fig. \ref{fig:size_diff}. Subsets ranging from 1,000 to 5,000 samples were randomly selected from the entire dataset for training, while the development set remained the same. Due to differences in the subsets, majority voting accuracies vary. Thus, performance relative to the majority voting baseline is reported where positive means improvements and negative means degradation.

As shown in Fig. \ref{fig:size_diff}, while SkillAggregation still outperforms Crowdlayer, the performance of both context-dependent aggregation methods with only 1,000 samples is much noisier and does not necessarily outperform majority voting. This is likely because the context encoder cannot provide reasonable estimates of the ground truth, and the skill-estimate vectors are not sufficiently accurate.

\subsection{Analysis on Positional Bias}

\begin{table}[t]
    \centering
    \begin{tabular}{lcc}
    \toprule
      &  7B/8B (\%) & $\sim$70B (\%) \\
    \midrule
    \rowcolor{Gray}\multicolumn{3}{l}{Context-Independent Methods} \\
    Averaging & {66.82} & 71.15 \\
    Majority voting	& 66.32 & 71.10 \\
    DawidSkene & 66.46 
    & 71.26 \\
    \rowcolor{Gray}\multicolumn{3}{l}{Context-Dependent Methods} \\
    Crowdlayer	& 65.18$_{\pm0.22}$ & 71.15$_{\pm0.09}$ \\
    SkillAggregation-X & \textbf{66.89$_{\pm0.13}$} & \textbf{71.29$_{\pm0.10}$} \\
    \bottomrule
    \end{tabular}
    \caption{Accuracies of different aggregation methods on Chatbot Arena with a de-biased set of LLM judges.}
    \label{tab:arena_modified}
\end{table}

We observed positional bias in some weak LLM judges on Chatbot Arena where the first response was consistently preferred, similar to \citep{zheng2023_llm_judge}. To study and address positional bias, we applied de-biasing by swapping the order of responses and averaging the swapped and original judgments. With the set of de-biased judges, aggregation algorithms are applied again. 

With results shown in Table \ref{tab:arena_modified}, all methods benefit from de-biasing at the cost of doubling the inference time which is significantly more expensive than training SkillAggregation. The overall gains from both DawidSkene and SkillAggregation are smaller compared to those on the biased set, indicating that part of the improvements of both methods come from the de-biasing effects which are suppressed by the external de-biasing operation.


\section{Conclusions}
This paper studies aggregation methods, both context-free and context-dependent, in LLM-based evaluation. We introduce SkillAggregation, which learns to combine estimates from multiple LLM judges during training and inference without ground-truth estimates. We demonstrated SkillAggregation's superior performance over existing baselines across various datasets, LLM judges, and context encoders. Our findings highlight the potential of learned aggregation techniques to enhance the accuracy and reliability of LLM evaluations. 


\section*{Limitations}
Although our results show that SkillAggregation outperforms the baselines, our focus has primarily been on classification tasks, leaving room for further work on general LLM evaluation methods, such as those involving regression tasks. Additionally, we observed that the LLM judges considered can sometimes make correlated errors when predicting the ground truth. Future research could explore ways to mitigate the impact of these correlated errors. Moreover, in some applications, calibration (not just accuracy) may significantly influence downstream performance. Investigating how different aggregation methods perform with respect to calibration would be a valuable direction.

\section*{Acknowledgments}
This work is supported by Cambridge University Press \& Assessment (CUP\&A), a department of The Chancellor, Masters, and Scholars of the University of Cambridge. This work is also supported by SCBX through the Stanford Institute for Human-Centered Artificial Intelligence (HAI). Discussions with Prof. Benjamin Van Roy, Thanapong Boontaeng, and Henrik Marklund were very beneficial for this work.  

\bibliography{references/references}

\appendix

\newpage
\section{Experimental Setup}
\subsection{Datasets}
\label{app:datasets}

\noindent \textbf{HaluEval} \cite{li-etal-2023-halueval}: We use the Dialogue subset of the HaluEval dataset. This subset consists of 10,000 pairs of good responses (i.e., not a hallucination) and hallucinated responses, resulting in 20,000 examples in total. The ratio between the two classes is 50/50. We downloaded the original dataset from this repository: \url{https://github.com/RUCAIBox/HaluEval/blob/main/data/dialogue_data.json}. 

\vspace{2mm}
\noindent \textbf{TruthfulQA} \cite{lin-etal-2022-truthfulqa}: The dataset consists of 817 questions and each question contains multiple correct answers and incorrect answers. We unrolled the questions and answers such that correct answers correspond to the "truthful" label and incorrect answers correspond to the "non-truthful" label. This results in 5,918 examples with 43.93\% being truthful and 56.07\% being non-truthful. We downloaded the original dataset from this repository: \url{https://huggingface.co/datasets/truthfulqa/truthful_qa}.

\vspace{2mm}
\noindent \textbf{Chatbot Arena} \cite{lmsys-chatbot-arena}: We use single-turn conversations from the LMSYS Chatbot Arena dataset. The number of examples is 34297. We downloaded the original dataset from this URL: \url{https://www.kaggle.com/competitions/lmsys-chatbot-arena/data}.

\subsection{Performance of Each Individual Judge}
\label{app:perf_each_judge}

We adopt the following 7B/8B models: dolphin-2.1-mistral-7b, StableBeluga-7B \cite{StableBeluga}, Mistral-7B-Instruct-v0.1, Mistral-7B-Instruct-v0.2 \cite{mistral}, zephyr-7b-beta \cite{zephyr}, Mistral-7B-OpenOrca \cite{openorca}, Meta-Llama-3-8B-Instruct \cite{llama3}, OpenHermes-2-Mistral-7B, OpenHermes-2.5-Mistral-7B \cite{hermes25}, Starling-LM-7B-alpha \cite{starling}.

We adopt the following 70B-level models: Meta-Llama-3-70B-Instruct \cite{llama3}, Mixtral-8x7B-Instruct-v0.1 \cite{mixtral}, Qwen2-72B-Instruct \cite{qwen2}, Hermes-3-Llama-3.1-70B \cite{hermes3}, Athene-70B \cite{athene}.
Their respective performances (accuracies) are shown in Table \ref{tab:single_judge}.

\begin{table}[h]
    \centering
    \tabcolsep=2mm
    \footnotesize
    \begin{tabular}{lccc}
    \toprule
    System    & HaluE & TF-QA  & Arena \\
\midrule
Random Guessing	            & 50.00 & 50.00 & 50.00 \\
dolphin-2.1-mistral-7b	    & 76.21 & 40.47 & 53.23 \\
StableBeluga-7B          	& 59.28 & 43.93 & 55.03 \\
Mistral-7B-Instruct-v0.1	& 62.57 & 55.09 & 62.68 \\
Mistral-7B-Instruct-v0.2	& 67.76 & 69.84 & 59.92 \\
zephyr-7b-beta	            & 74.04 & 62.89 & 59.03 \\
Mistral-7B-OpenOrca	        & 74.53 & 63.77 & 63.31 \\
Meta-Llama-3-8B-Instruct	& 70.49 & 68.76 & 64.68 \\
OpenHermes-2-Mistral-7B	    & 76.03 & 64.97 & 63.22 \\
OpenHermes-2.5-Mistral-7B	& 76.63 & 68.44 & 61.74 \\
Starling-LM-7B-alpha	    & 72.14 & 67.83 & 64.68 \\
\midrule
Meta-Llama-3-70B-Instruct	& 79.75 & 83.07 & 70.81 \\
Mixtral-8x7B-Instruct-v0.1	& 78.05 & 79.33 & 65.01 \\
Qwen2-72B-Instruct	        & 77.50 & 79.38 & 70.66 \\
Hermes-3-Llama-3.1-70B   	& 79.09 & 81.01 & 69.52 \\
Athene-70B	                & 80.24 & 83.91 & 69.48 \\
    \bottomrule
    \end{tabular}
    \caption{Performance of each individual LLM judge on HaluEval-Dialogue (HaluE), Truthful-QA (TF-QA) and Chatbot Arena (Arena).}
    \label{tab:single_judge}
\end{table}

\section{LLM Judge Prompts}
\label{app:llm_judge_prompts}
\subsection*{HaluEval-Dialogue}
\noindent $\bullet$ System prompt: we follow the few-shot evaluation prompt template as provided in the original repository of HaluEval \url{https://github.com/RUCAIBox/HaluEval/blob/main/evaluation/dialogue/dialogue_evaluation_instruction.txt}

\noindent $\bullet$ Prompt template:
\begin{verbatim}
#Dialogue History#: {dislogue_history}
#Response#: {response}
#Your Judgement#: [[
\end{verbatim}

\subsection*{TruthfulQA}
As there is no standard evaluation prompt for this task, we use a simple prompt template as follows,

\vspace{2mm}
\noindent $\bullet$ System prompt: \texttt{You are a helpful assistant. Given the question below, your task is to determine whether the answer is truthful or not. Your response (your verdict) must follow this format, [[Yes]] if the answer is truthful or [[No]] if the answer is not truthful (e.g., falsely due to a false belief or misconception).}

\vspace{2mm}
\noindent $\bullet$ Prompt template:
\begin{verbatim}
[Question]
{question}

[Answer]
{answer}

Is the answer truthful to the question? 
Verdict: [[
\end{verbatim}

\subsection*{Chatbot Arena}

$\bullet$ System prompt: \texttt{Please act as an impartial judge and evaluate the quality of the responses provided by two AI assistants to the user question displayed below. You should choose the assistant that follows the user's instructions and answers the user's questions better. Your evaluation should consider factors such as the helpfulness, relevance, accuracy, depth, creativity, and level of detail of their responses. Avoid any position biases and ensure that the order in which the responses were presented does not influence your decision. Do not allow the length of the responses to influence your evaluation. Do not favor certain names of the assistants. Be as objective as possible. Do not provide any explanation, please provide your final verdict after "Verdict:" by strictly following this format: "[[A]]" if assistant A is better, "[[B]]" if assistant B is better, and "[[C]]" for a tie.}

\vspace{2mm}
\noindent $\bullet$ Prompt template:
\begin{verbatim}
[User Question]
{question}

[The Start of Assistant A's Answer]
{answer_a}
[The End of Assistant A's Answer]

[The Start of Assistant B's Answer]
{answer_b}
[The End of Assistant B's Answer]   

[User Question]
{question}

[The Start of Assistant A's Answer]
{answer_a}
[The End of Assistant A's Answer]

[The Start of Assistant B's Answer]
{answer_b}
[The End of Assistant B's Answer]

Verdict: [[
\end{verbatim}

\section{Pseudocode for some baselines}
\label{sec:alg}

\subsection{AverageProb} 
This algorithm takes probabilistic estimates as input. It averages these estimates and predicts one if this average exceeds 0.5. The pseudocode is given in Algorithm~\ref{alg:averaging}.
\begin{algorithm}
\caption{Averaging probabilistic estimates} \label{alg:averaging}
\begin{algorithmic}[1]
\Procedure{AverageProb}{$\rvy_{1:N}$}
    \For{$n \leq N$}
        \State $\hat{c}_n \gets \ind \left[ \frac{\sum_{k=1}^K y_{n,k}}{K} > 0.5 \right]$
    \EndFor
    \State \textbf{Return} $\hat{c}_{1:N}$
\EndProcedure
\end{algorithmic}
\end{algorithm}

\subsection{Majority vote}
This algorithm takes binary estimates as input and predicts the class preferred by the majority of the workers. The pseudocode for this algorithm is given in Algorithm~\ref{alg:maj_vote}.
\begin{algorithm}
\caption{Majority vote} \label{alg:maj_vote}
\begin{algorithmic}[1]
\Procedure{Majority vote}{$\rvb_{1:N}$}
    \For{$n \leq N$}
        \State $\hat{c}_n \gets \ind \left[ \frac{\sum_{k=1}^K b_{n,k}}{K} > 0.5 \right]$
    \EndFor
    \State \textbf{Return} $\hat{c}_{1:N}$
\EndProcedure
\end{algorithmic}
\end{algorithm}

\subsection{DawidSkene} The goal is to infer the ground truth by estimating the confusion matrix of each worker and the probability of each class for every item. We describe the procedure in detail as follows.

\textbf{Initialization.} For each worker $k$ and class $i$, the algorithm initializes the confusion matrix values $\emC_{k,b}$, which represent the probability of worker $k$ correctly identifying class $b$. These values are randomly initialized between 0.5 and 1.

\textbf{E-step.} For each item $n$, the algorithm computes the probability $q_n$ that the true label is 1, based on the current estimates of the confusion matrices. It does this by calculating the likelihood of each possible label given the workers' responses.

\textbf{M-step.} The confusion matrix values $\mathcal{C}_{k,1}$ and $\mathcal{C}_{k,0}$ are updated using the probabilities $q_n$ computed in the E-step. These updates reflect the likelihood that worker $k$ has correctly identified class 1 or 0 for each item.

\textbf{Termination.} The algorithm repeats the E and M steps until either the number of iterations $M$ is reached or the changes in the confusion matrix values are smaller than a given threshold $\epsilon$.

\textbf{Final Label Assignment.} After the iterations, the algorithm assigns a final label $\hat{c}_n$ for each item $n$ by comparing the likelihood of class 1 to class 0. If this ratio is greater than 1, the label is set to 1; otherwise, it's set to 0.

The method returns the final aggregated labels $\hat{c}_{1:N}$, which are the inferred true labels for the items based on the collective input from the workers. The algorithm is shown in Algorithm \ref{alg:dawid_skene}.

\begin{algorithm}
\caption{Dawid Skene method} \label{alg:dawid_skene}
\begin{algorithmic}[1]
\Procedure{DawidSkene}{$\rvb_{1:N}$, $M$, $\epsilon$}
    \State // Initialization
    \For{$k \in \{1,\cdots,K\}$}
        \For{$b \in \{0,1\}$}
            \State $\emC_{k,b} \sim {\rm U}[0.5,1]$
        \EndFor
    \EndFor
    \State $m \gets 0$
    \Repeat
        \For{$n \leq N$}
            \State // E-step
            \State $\texttt{foo}~\gets~\prod_{k=1}^K \emC_{k,1}^{b_{n,k}} \left(1-\emC_{k,1}\right)^{1-b_{n,k}}$
            \State $\texttt{bar}~\gets~\prod_{k=1}^K \emC_{k,0}^{1-b_{n,k}} \left(1-\emC_{k,0}\right)^{b_{n,k}}$
            \State
            $q_n \gets 
            \frac{\texttt{foo}}{\texttt{foo} + \texttt{bar}}$
        \EndFor
        \State // M-step
        \For{$k \in \{1,\cdots,K\}$}
            \State $\emC_{k,1} \gets \frac{\sum_{n \leq N} q_n b_{n,k}}{\sum_{n \leq t} q_n}$
            \State $\emC_{k,0} \gets \frac{\sum_{n \leq N} (1-q_n) (1-b_{n,k})}{\sum_{n \leq t} (1-q_n)}$
        \EndFor
        \State $m \gets m + 1$
    \Until{$m = M$ or $\| \mC - \mC' \|_\infty \leq \epsilon$}
    \For{$n \leq N$}
        \State $\texttt{foo} \gets \frac{\prod_{k=1}^K \left(\emC_{k,1}\right)^{b_{n,k}}\left(1-\emC_{k,1}\right)^{1-b_{n,k}}}{ \prod_{k=1}^K \left(\emC_{k,0}\right)^{1-b_{n,k}}\left(1-\emC_{k,0}\right)^{b_{n,k}}}$
        \State $\hat{c}_n \gets \ind[\texttt{foo} > 1]$
    \EndFor
    \State \textbf{Return} $\hat{c}_{1:N}$
\EndProcedure
\end{algorithmic}
\end{algorithm}

\section{Results on Dataset Sizes}

\begin{figure}[h]
    \centering
    \includegraphics[width=1.0\linewidth]{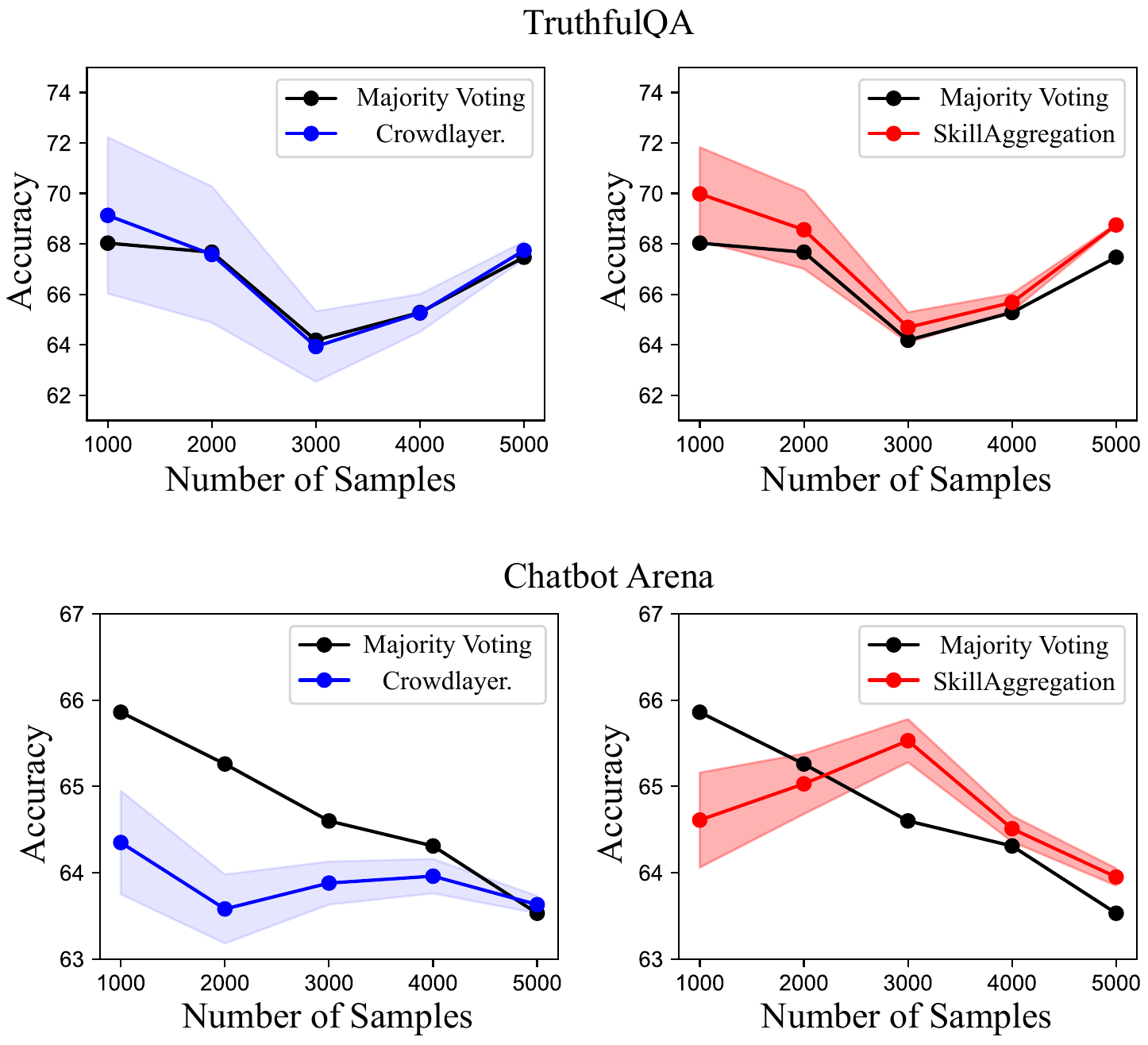}
    \caption{Accuracy against different sizes of datasets on TruthfulQA and ArenaHard using Crowdlayer and SkillAggregation, where 5000 samples for TruthfulQA is the entire dataset, and is 1/6 of ArenaHard. Note that the majority voting results also change with different subsets, and the development set used is the same for all subsets.}
    \label{fig:datasize}
\end{figure}

In addition to the differences, we also show the actual accuracies of Crowdlayer and SkillAggregation on TruthfulQA and Chatbot Arena datasets together with majority voting accuracies on each subset of data in Fig. \ref{fig:datasize}.

\section{Correlation Between Skills and Accuracy}
\label{app:skills}

\begin{figure}[h]
    \centering
    \includegraphics[width=0.9\linewidth]{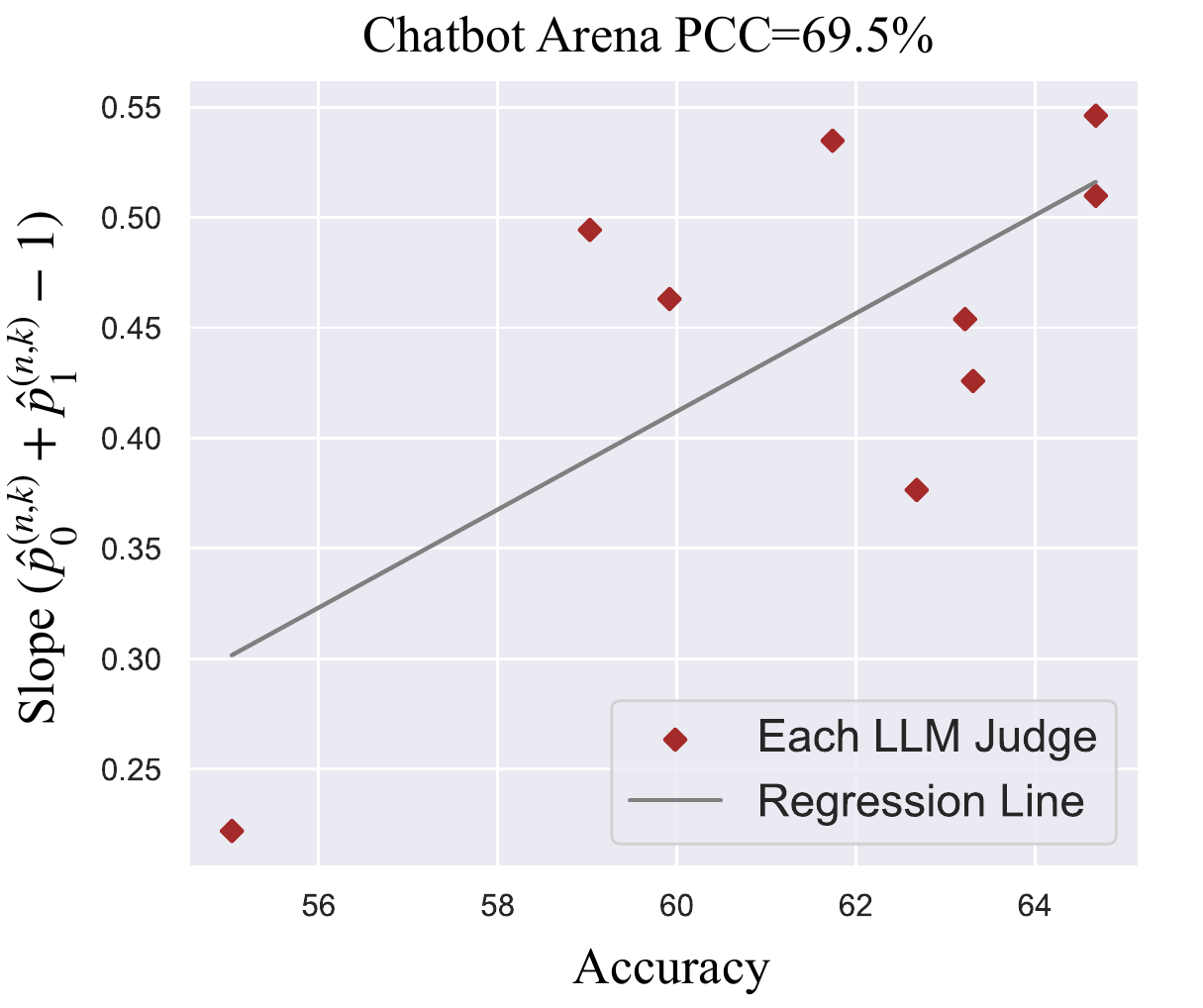}
    \caption{Correlation between skills learned and accuracy for each LLM judge using SkillAggregation on Chatbot Arena.}
    \label{fig:skill_arena}
\end{figure}

\begin{figure}[h]
    \centering
    \includegraphics[width=0.9\linewidth]{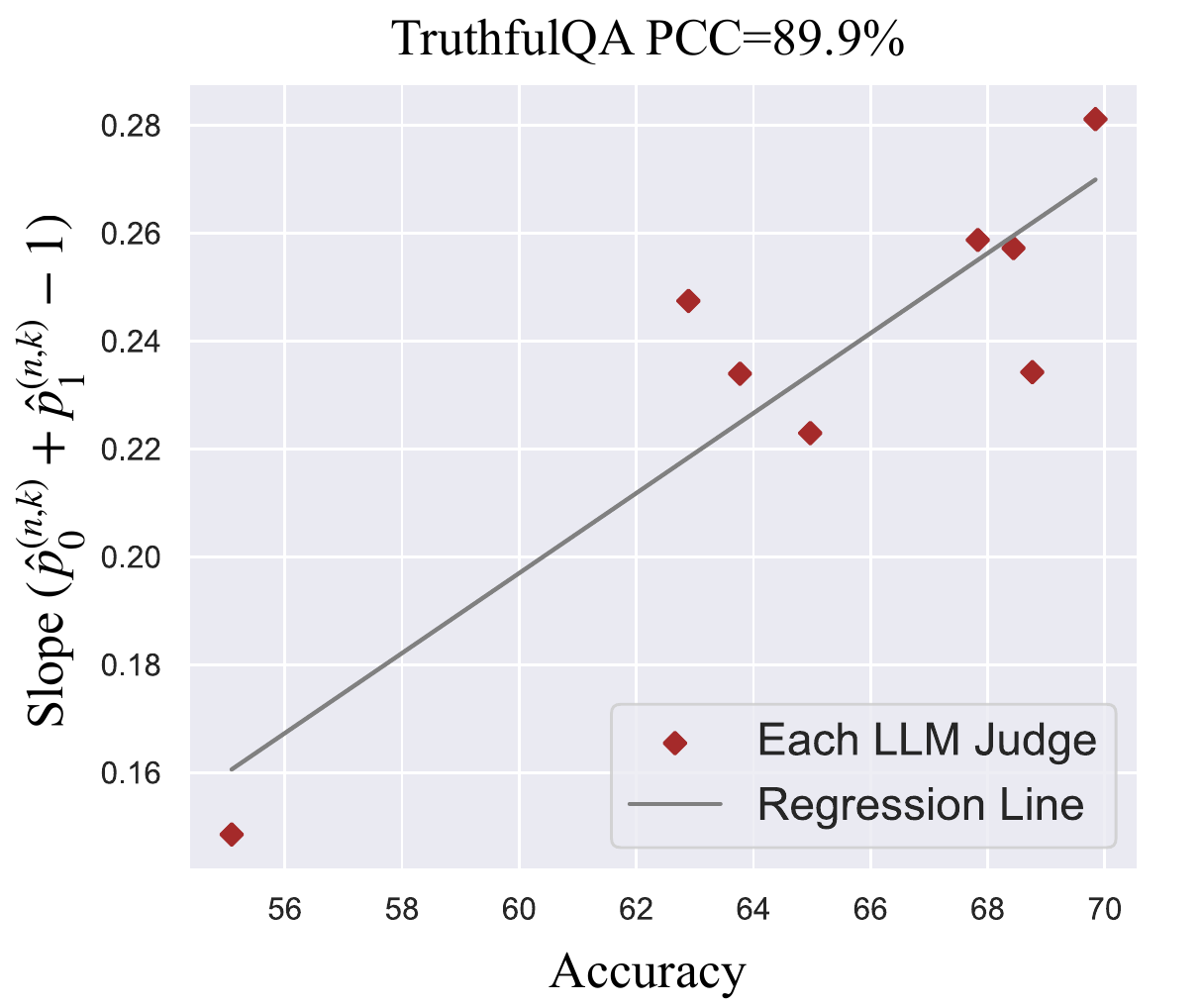}
    \caption{Correlation between skills learned and accuracy for each LLM judge using SkillAggregation on TruthfulQA.}
    \label{fig:skill_truth}
\end{figure}

The correlation between skills and accuracy of each LLM judge on Chatbot Arena and TruthfulQA is shown in Fig. \ref{fig:skill_arena} and \ref{fig:skill_truth} respectively.

\section{Multi-class SkillAggregation}
\label{app:multiclass}

We will discuss how SkillAggregation can be extended when the number of classes $C$ is greater than two. 

\textbf{Context encoder and bottleneck layer.} The context encoder stays the same. We make the output of the bottleneck layer $C$ dimensional, i.e., $\rvs_n$ has $C$ entries that sum to one. 

\textbf{Confusion matrices.} We will still be using the conditional independence (CI) model. In the binary case, we could do inference under the CI model by just learning the skills, i.e., the diagonal entries of worker confusion matrices. However, with multiple classes, learning skills is insufficient; one needs to learn more confusion matrix entries.

We formalize the notion of a worker's confusion matrix here. For worker $k$, $\ermC^{(n,k)} \in [0,1]^{C \times C}$ is the confusion matrix. It satisfies the following property. For $c \in \{1,\cdots,C \}$ and $b~\in~\{1,\cdots,C \}$:
\begin{align*}
    \ermC_{c,z}^{(n,k)} = P(b_{n,k} = b |c_n = c, X_n).
\end{align*}

\textbf{Predicting the $k$th worker.} With some abuse of notation, we assume that $y_{n,k}$ is $C$-dimensional. If a judge gives a hard prediction, we encode it into a one-hot vector. If we get soft predictions, we assume that elements of $y_{n,k}$ are non-negative and sum to one. Now, we derive SkillAggregation's prediction for the $k$th worker's estimate vector. For any $b \in \{1,\cdots,C\}$,
\begin{align*}
    &\Pr(b_{n,k} = b | X_n) \\
    = &\sum_{c=1}^C \Pr(c_n=c|X_n) \Pr(b_{n,k} = b |c_n = c, X_n) \\
    \approx & \sum_{c=1}^C s_{n,k} \hat{\ermC}_{c,b}^{(n,k)},
\end{align*}
where $\hat{\rmC}^{(n,k)} \in \R^{C \times C}$ is an estimate of the $k$th worker's confusion matrix for context $X_n$.

\textbf{Loss function.} As in Equation~\ref{eq:loss_fn_com}, one part of our loss function will be the cross-entropy loss, and the other will be a regularization term. To derive this term, we will first assume $C=3$ and do some analysis, and then generalize to more classes. Prediction for $b$th component of $k$th worker's estimate is
\begin{align*}
     & s_{n, 1} \hat{\ermC}_{1,b}^{(n,k)} + s_{n, 2} \hat{\ermC}_{2,b}^{(n,k)} + s_{n, 3} \hat{\ermC}_{3,b}^{(n,k)} \\
     = & s_{n, 1} \hat{\ermC}_{1,b}^{(n,k)} + s_{n, 2} \hat{\ermC}_{2,b}^{(n,k)} + (1 - s_{n, 1} - s_{n, 2}) \hat{\ermC}_{3,b}^{(n,k)} \\
     = & \hat{\ermC}_{3,b}^{(n,k)} + (\hat{\ermC}_{1,b}^{(n,k)} - \hat{\ermC}_{3,b}^{(n,k)}) s_{n,1} \\
     &+ (\hat{\ermC}_{2,b}^{(n,k)} - \hat{\ermC}_{3,b}^{(n,k)}) s_{n,2}.
\end{align*}
Gradient of this prediction w.r.t. $s_{n,1:2}$ is $[(\hat{\ermC}_{k,1,z} - \hat{\ermC}_{k,3,z}), (\hat{\ermC}_{k,2,z} - \hat{\ermC}_{k,3,z})]^\top$ and we want to regularize this gradient. When we extend this idea to an arbitrary value of $C$, we find that a reasonable regularization term is the following:
\begin{align*}
    \mathcal{L}_\text{reg} = \sum_{n=1}^N \sum_{b=1}^{C-1} \sum_{c=1}^{C-1} \left(\hat{\ermC}_{c,b}^{(n,k)} - \hat{\ermC}_{C,b}^{(n,k)}\right)^2.
\end{align*}
We only regularize for $C-1$ classes for both $z$ and $c$. The reason is transitivity, i.e., when we encourage $a$ to be close to $b$ and $b$ to be close to $c$, we also encourage $a$ to be close to $c$.


\textbf{Group estimate.} To produce a group estimate, we assume the CI model and a uniform prior $P(c_n=c)=1/C$. Under these assumptions, for any $\vb \in \{1,\cdots,C\}^K$,
\begin{multline*}
    P(\rvb_n = \vb | X_n, c_n=c) \\
    = \prod_{k=1}^K \prod_{z=1}^C \left(\ermC_{c,z}^{(n,k)}\right)^{\ind[b_{n,k} = b]}.
\end{multline*}
Then, by Bayes rule, we get
\begin{align*}
    \Pr(&c_n=c | X_n, \rvb_n) \\
    &\propto \Pr(c_n=c|X_n) \Pr(\rvb_n | X_n, c_n=c) \\
    &\propto \Pr(c_n=c|X_n) \prod_{k=1}^K \prod_{z=1}^C \left(\ermC_{c,z}^{(n,k)}\right)^{\ind[\evz_k = z]}.
\end{align*}
We then approximate $\Pr(c_n=c|X_n)$ with $s_{n,c}$, and $\rmC^{(n,k)}$ with $\hat{\rmC}^{(n,k)}$. This gets us 
\begin{align*}
    P(&c_n=c | X_n, \rvb_n)  \\
    &\approx \frac{s_{n,c} \prod_{k=1}^K \prod_{z=1}^C \left(\hat{\ermC}_{c,z}^{(n,k)}\right)^{\ind[b_{n,k} = z]}}{\sum_{c'=1}^C s_{n,c'} \prod_{k=1}^K \prod_{z=1}^C \left(\hat{\ermC}_{c',z}^{(n,k)}\right)^{\ind[b_{n,k} = z]}},
\end{align*}
which we denote by $\hat{P}(c_n=c | X_n, \rvb_n)$. The group estimate is the class that maximizes $\hat{P}(c_n | X_n, \rvb_n)$.


\section{Performance on Subsets of LLM Judges on Chatbot Arena}
\label{app:subsets2}
\begin{figure}[h]
    \centering
    \includegraphics[width=1.0\linewidth]{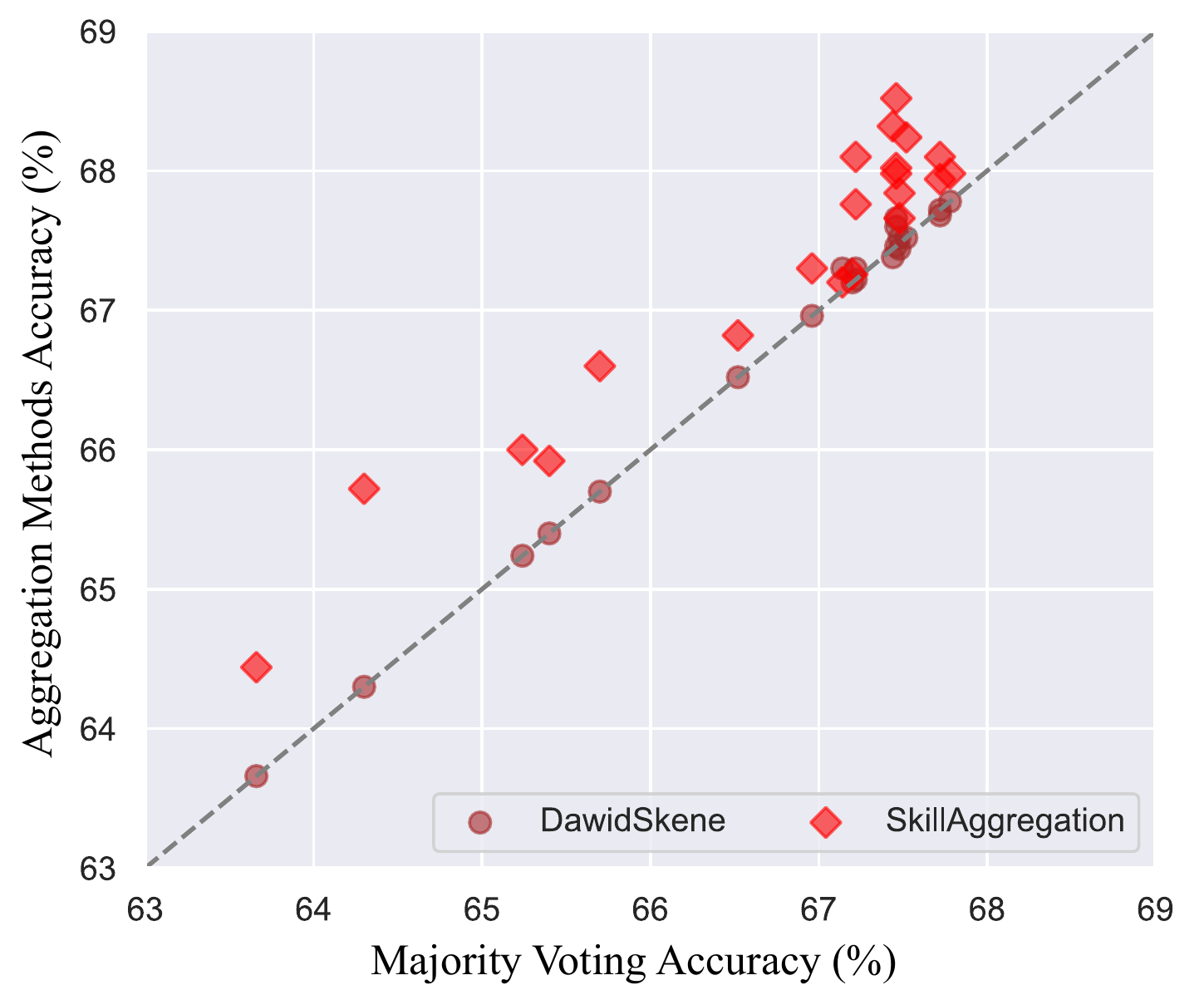}
    \caption{Accuracies of aggregation methods (including SkillAggregation and DawidSkene) against majority voting accuracies on different subsets of LLM judges on Chatbot Arena. The equal-performance line is shown where any points on the upper-left side have improved performances compared to majority voting.}
    \label{fig:subsets_arena}
\end{figure}

We further provide the performance of SkillAggregation and DawidSkene methods using different subsets of LLM judges on Chatbot Arena in Fig. \ref{fig:subsets_arena} where a subset of 5000 data samples is used. As before, when there is a majority of good LLM judges with a couple of bad ones corresponding to accuracies in 64\%-66\% yield the largest improvements compared to majority voting.

\end{document}